\documentclass[12pt]{article}
\usepackage{amssymb,epsfig,psfrag,amsmath,art,cite}
\newfont{\smalll}{cmr8}

\def\IR{{\mbox{\hbox{I\hskip-.1em R}}}}

\def\IC{\hbox{C\hskip-
.5em\raise.5ex\hbox{$\scriptscriptstyle\mid$}}\ }
\def\Ic{\hbox{\smalll C\hskip-
.5em\raise.3ex\hbox{$\scriptscriptstyle\mid$}}\ }

\def\T={\buildrel {\scriptscriptstyle\triangle} \over =}

\def\sqr#1#2{{\vcenter{\vbox{\hrule height.#2pt\hbox{\vrule
width.#2pt height#1pt \kern#1pt\vrule width.#2pt}\hrule
height.#2pt}}}}
\def\square{\mathchoice\sqr64\sqr64\sqr33\sqr33}

\def\block-diag{\mathop{\rm block{\scriptstyle -}diag}}

\def\pmbb#1{\setbox0=\hbox{#1}\raise 0.5ex\box0}

\def\l{\lambda}

\def\nxn{{n\times n}}

\def\T{^{\rm T}}

\newtheorem{theorem}{Theorem}[section]

\newtheorem{remark}{Remark}[section]
\newtheorem{definition}{Definition}[section]

\newtheorem{problem}{Problem}[section]

\newcommand{\defeq}{\stackrel{\triangle}{=}}

\def\IR{{\mathbb R}}
\def\IC{{\mathbb C}}

\renewcommand\theenumi{\roman{enumi}}
\renewcommand\theenumii{\alph{enumii}}
\renewcommand\theenumiii{\arabic{enumiii}}
\renewcommand\theenumiv{\Alph{enumiv}}

\def\sc{s_{\rm c}}

\newcommand{\ZA}{Z_\mathcal{A}}

\begin{document}

\title{Segmentation of Facial Expressions Using Semi-Definite Programming and Generalized\\ Principal Component Analysis}

%{\Large\bf ACC02-IEEE1522}\\

\date{August 2008}

\def\bgat{\normalsize\begin{tabular}{c c c}
Behnood Gholami & & Allen R. Tannenbaum\\
School of Aerospace Engineering & & Schools of Electrical \& Computer and \\
Georgia Institute of Technology & & Biomedical Engineering\\
behnood@gatech.edu & & Georgia Institute of Technology\\
&& tannenba@ece.gatech.edu\\
\end{tabular}}

\def\wmh{\normalsize\begin{tabular}{c}
Wassim M. Haddad\\
School of Aerospace Engineering\\
Georgia Institute of Technology \\
wm.haddad@aerospace.gatech.edu\\
\end{tabular}}

\author{\bgat \\ \\ \wmh \\}

\maketitle

\baselineskip 14pt
\begin{abstract}
In this paper, we use semi-definite programming and generalized principal component analysis (GPCA) to distinguish between two or more different facial expressions. In the first step, semi-definite programming is used to reduce the dimension of the image data and ``unfold'' the manifold which the data points (corresponding to facial expressions) reside on. Next, GPCA is used to fit a series of subspaces to the data points and associate each data point with a subspace. Data points that belong to the same subspace are claimed to belong to the same facial expression category. An example is provided.
  \vfill
\end{abstract}
%\vfill

\newpage

%%%%%%%%%%%%%%%%%%%%%%%%%%%%%%%%%%%%%%%%%%%%%%%%%%%%%%%%%%%%%%%%%%%%%%%%%%%%%%%%%%%%%%%%%%%%%%%%%%%%%%%%%%%%%%%%%%%%%%%%%%%%%
\section{Mathematical Preliminaries}

In this section, we introduce notation, several
definitions, and some key results in abstract algebra and algebraic geometry \cite{Eisenbud:1995,Tannenbaum:1981,Gallian:1990,VMS:2008} that are
necessary for developing the main results of this paper.
Specifically, for $A\in\IR^\nxn$ we write $A\ge0$ (resp.,
$A>0$) to indicate that $A$ is a nonnegative-definite (resp.,
positive definite) matrix. In addition, $(\cdot)^{\rm T}$ denotes transpose, and $(\cdot)^\dag$
denotes the Moore-Penrose generalized inverse. In the next paragraphs we give the definitions for ideal and the Veronese map.

\begin{definition}[\hspace*{-.1em}\textbf{(Ideal)}]
  Let $\mathcal{R}$ be a commutative ring and $I$ be an additive subgroup of $\mathcal{R}$. $I$ is called an ideal if $r\in\mathcal{R}$ and $s\in I$ then $rs\in I$. Furthermore, an ideal is said to be generated by a set $S$, if for all $t\in I$, $t=\sum_{i=1}^{n}r_is_i$, $r_i\in\mathcal{R}$, $s_i\in S$, $i=1,2,\dots n$ for some $n\in\mathbb{N}$.

\end{definition}

Let $R[x]$ be the set of polynomials of $D$ variables, where $x=[x_1\, x_2\, \dots \, x_D]\T$, $x_i\in R$, $i=1,2,\dots D$, and $R$ is a field. Then $R[x]$ with the polynomial addition and multiplication is a commutative ring. A product of $n$ variables $x_1,x_2,\dots x_n$ is called a monomial of degree $n$ (counting repeats). The number of distinct monomials of degree $n$ is given by
\begin{eqnarray}
  M_n(D) \defeq \left(
             \begin{array}{c}
               D+n-1 \\
               n \\
             \end{array}
           \right).\label{eq:MnD}
\end{eqnarray}

\begin{definition}[\hspace*{-.3em}\textbf{(Veronese Map)}\cite{VMS:2008}]
The Veronese Map of degree $n$, $\nu_n:R^D\rightarrow R^{M_n(D)}$, is a mapping that assigns to $D$ variables $x_1,x_2,\dots x_D$, all the possible monomials of degree $n$, namely,
\[ \nu([x_1\, x_2\, \dots \, x_D]\T)=[u_1\, u_2\, \dots \, u_{M_n(D)}]\T\]
such that $u_i=x_1^{n_{i1}}x_2^{n_{i2}}\dots x_D^{n_{iD}}$, $i=1,2,\dots M_n(D)$, where $n_{i1}+n_{i2}+\dots+n_{iD}=n$, $n_{ij}\in \mathbb{N}$, $j=1,2,\dots D$, and $M_n(D)$ is given by (\ref{eq:MnD}).
\end{definition}

%%%%%%%%%%%%%%%%%%%%%%%%%%%%%%%%%%%%%%%%%%%%%%%%%%%%%%%%%%%%%%%%%%%%%%%%%%%%%%%%%%%%%%%%%%%%%%%%%%%%%%%%%%%%%%%%%%%%%%%%%%%%%
%%%%%%%%%%%%%%%%%%%%%%%%%%%%%%%%%%%%%%%%%%%%%%%%%%%%%%%%%%%%%%%%%%%%%%%%%%%%%%%%%%%%%%%%%%%%%%%%%%%%%%%%
\section{Dimension Reduction}\label{sec:dimred}

In this section, we introduce a method known as Maximum Variance Unfolding (MVU), a dimension reduction technique which uses semi-definite programming. Given a set of points sampled from a low dimensional manifold in a high dimensional ambient space, this technique ``unfolds'' the manifold (and hence, the points it contains) while preserving the local geometrical properties of the manifold \cite{WS:IJCV:2006}. This method, in a sense, can be regarded as a nonlinear generalization for the Principal Component Analysis (PCA). Given a set of points in a high dimensional ambient space, PCA identifies a low dimensional subspace such that the variance of the projection of the points on this subspace is maximized. More specifically, the base of a subspace on which the projection of the points has the maximum variance is the eigenvectors corresponding to the non-zero eigenvalues of the covariance matrix \cite{Jolliffe:2002}. In the case where data is noisy, the singular vectors corresponding to the dominant singular values of the covariance matrix are selected \cite{VMS:2008}.

Given $N$ input points $\{x_n\}_{n=1}^{N}\in\IR^D$, we would like to find $N$ output points $\{y_n\}_{n=1}^{N}\in\IR^d$ such that $d<D$, there is a one-to-one correspondence between these sets, and points \emph{close} to each other in the input data set remain \emph{close} in the output data set. In order to be more precise, we need to introduce the concept of \emph{isometry} for a set of points \cite{TSL:SCI:2000,WS:IJCV:2006}. Loosely speaking, isometry is an invertible smooth mapping defined on a manifold such that it locally has a translation and rotation effect. The next definition extends the notion of isometry to data sets.

\begin{definition}[\hspace*{-.3em}\cite{WS:IJCV:2006}]
  Let $X=\{x_n\}_{n=1}^{N}\in\IR^D$ and $Y=\{y_n\}_{n=1}^{N}\in\IR^d$ be two sets of point that are in one-to-one correspondence. Then $X$ and $Y$ are $k$-locally isometric if there exists a mapping consisting of rotation and translation $T:\IR^D\rightarrow\IR^d$ such that if $T(x_n)=y_n$ then $T(N_{x_n}(k))=N_{y_n}(k)$, for $n=1,2,\dots n$, where $N_x(k)$ is the set of $k$-nearest neighbors of $x \in X$.
\end{definition}

Before stating the MVU method, we give the problem statement.

\begin{problem}\label{prob:1}
  Given a set of input data points $X=\{x_n\}_{n=1}^{N}\in\IR^D$ find the output data points $Y=\{y_n\}_{n=1}^{N}\in\IR^d$, $d\le D$, such that the sum of pairwise square distances between outputs, namely,
   \begin{eqnarray}
     \Phi=\frac{1}{2n}\sum_{i=1}^N \sum_{j=1}^N \|y_i-y_j\|^2,\label{eq:1}
   \end{eqnarray}
   is maximized and $X$ and $Y$ are $k$-locally isometric for some $k\in\mathbb{N}$. Without loss of generality, we assume that $\sum_{n=1}^N x_n=0$. Moreover, we require $\sum_{n=1}^{N} y_n = 0$ to remove the translational degree of freedom of the output points $Y$.
\end{problem}

\begin{figure}[t] %h:here, t:top, b:bottom, p:newpage
   \hfil
   \begin{minipage}[h]{.35\linewidth}
      \centering
      \psfig{file=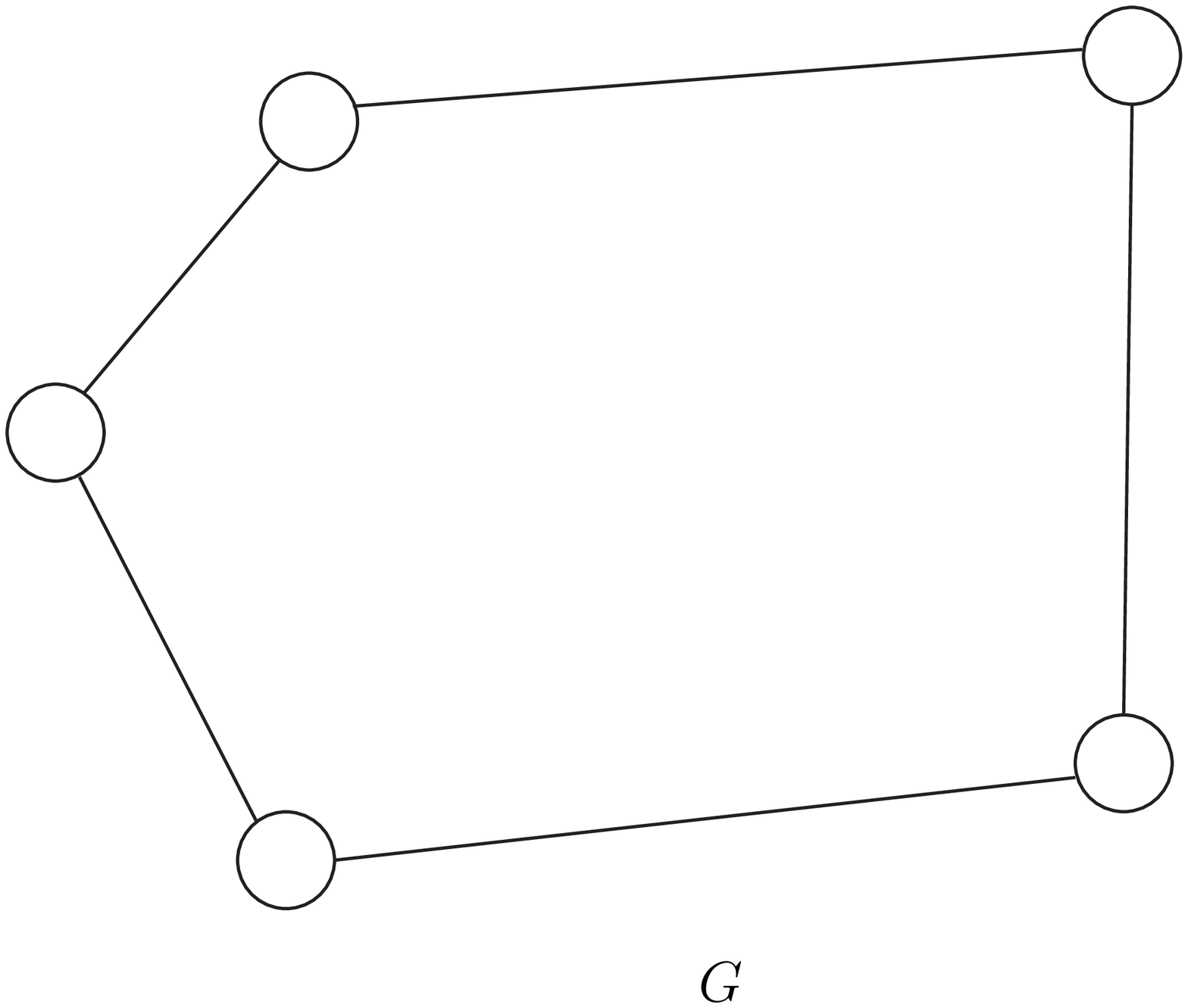,width=\linewidth}
   \end{minipage} \hfil
   \begin{minipage}[h]{.35\linewidth}
      \centering
      \psfig{file=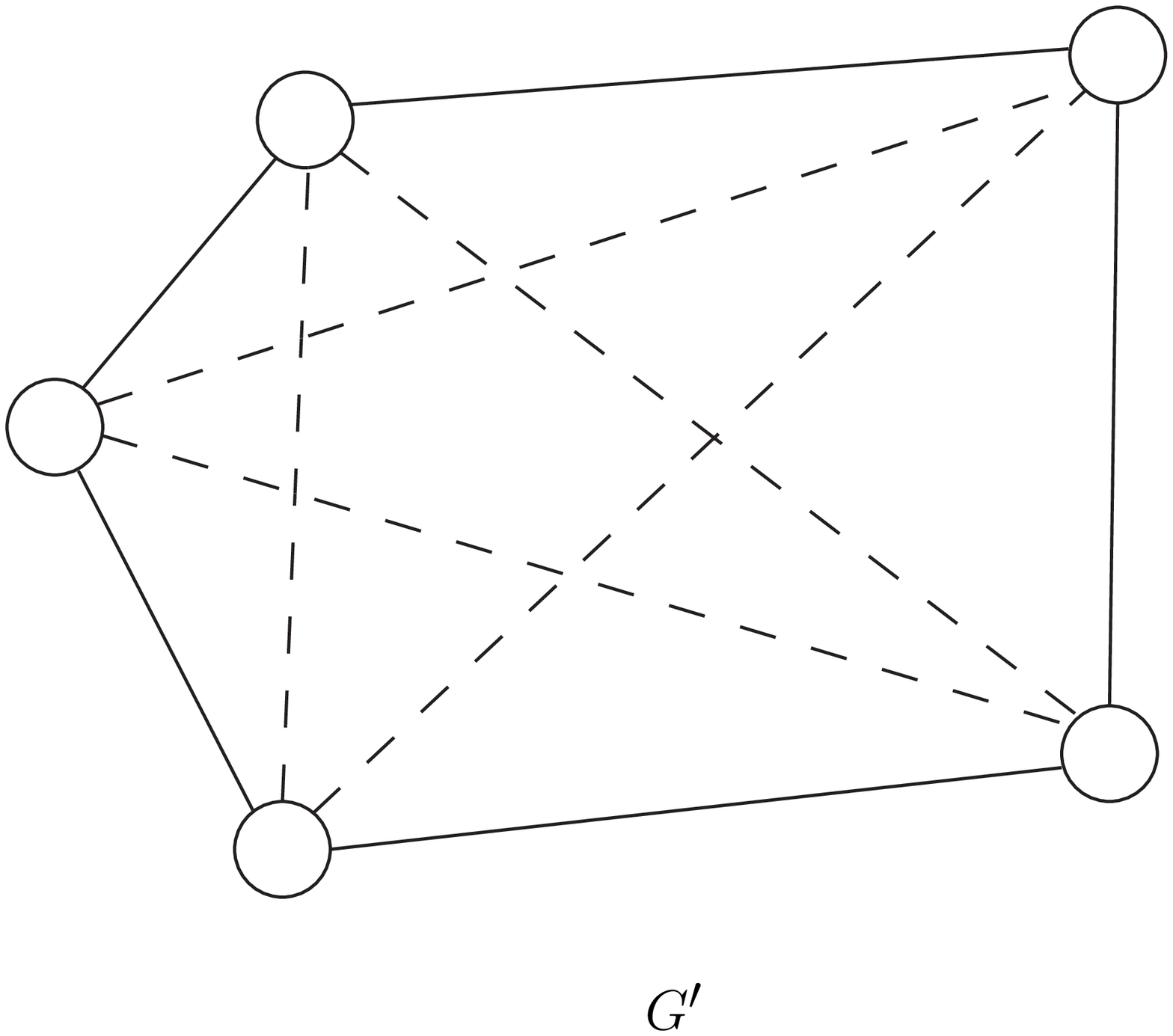,width=\linewidth}
   \end{minipage}
   \hfil
   \caption {The original and modified graphs for $k=2$}\label{fig:1}
\end{figure}

Note that the data set can be represented by a weighted graph $G$, where each node represents a point and the $k$-nearest points are connected by edges where $k$ is a given parameter. The weights also represent the distance between the nodes. We, furthermore, assume that the corresponding graph $G$ is connected. In case of a disconnected graph, each connected component should be analyzed separately. The $k$-local isometry condition in Problem \ref{prob:1} requires that the distances and the angles between the $k$-nearest neighbors to be preserved. This constraint is equivalent to merely preserving the distances between neighboring points in a modified graph $G'$, where in $G'$ for each node, all the neighboring nodes are also connected by an edge. More precisely, in $G'$, each node and the $k$-neighboring nodes form a clique of size $k+1$ (See Figure \ref{fig:1}).

The next theorem gives the solution to Problem \ref{prob:1} for the case $d=D$.
\begin{theorem}[\hspace*{-.3em}\cite{WS:IJCV:2006}]\label{thm:1}
  Consider the problem given by Problem \ref{prob:1} and assume $d=D$. The output data points $Y=\{y_n\}_{n=1}^{N}\in\IR^D$ are given by the solution to the optimization problem
  \begin{eqnarray}
    \max \Phi ,\label{eq:1.1}
  \end{eqnarray}
  subject to
  \begin{eqnarray}
    \sum_{n=1}^N y_n &=& 0,\label{eq:2}\\
    \|y_i-y_j\|^2 &=& D_{ij},\quad \textrm{if } \eta_{ij}=1,\quad i,j=1,2,\dots N,\label{eq:3}
  \end{eqnarray}
where $\Phi$ is defined in (\ref{eq:1}), $\eta=[\eta_{ij}]\in\IR^{N\times N}$ is the adjacency matrix of the modified graph $G'$, and $D_{ij}=\|x_i-x_j\|^2$, $i,j=1,2,\dots N$, $x_i,x_j\in X$.
\end{theorem}

The optimization problem (\ref{eq:1.1})--(\ref{eq:3}) is \emph{not} convex. The following convex optimization problem, however, is equivalent to the optimization problem given in Theorem \ref{thm:1}. Moreover, this theorem also addresses the case where $d\le D$.

\begin{theorem}[\hspace*{-.3em}\cite{WS:IJCV:2006}]\label{thm:2}
  Consider the problem given by Problem \ref{prob:1} and assume that $d=D$. The output data points $Y=\{y_n\}_{n=1}^{N}\in\IR^D$ are given by the solution to the optimization problem
  \begin{eqnarray}
    \max {\rm tr} (K) ,\label{eq:4}
  \end{eqnarray}
  subject to
  \begin{eqnarray}
    K &\ge & 0,\label{eq:4.9}\\
    \sum_{i=1}^N\sum_{j=1}^N K_{ij}&=&0,\label{eq:5}\\
    K_{ii}-2K_{ij}+K_{jj}&=&D_{ij},\quad \textrm{if } \eta_{ij}=1,\quad i,j=1,2,\dots N,\label{eq:6}
  \end{eqnarray}
where $K=[k_{ij}]$ is the inner product matrix defined by $k_{ij}=y_i\T y_j$, $i,j=1,2,\dots N$, and $\eta$ and $D_{ij}$ are defined in Theorem \ref{thm:1}. Moreover,
\begin{eqnarray}
  y_{ni}=\sqrt{\l_n V_{ni}},\quad i=1,2,\dots D,\quad n=1,2,\dots N,
\end{eqnarray}
where $V_n=[V_{n1} V_{n2} \dots V_{nD}]\T$, $n=1,2,\dots N$, is the eigenvector of $K$, $\l_n$ is its associated eigenvalue, and $y_n=[y_{n1}\, y_{n2}\, \dots \, y_{nD}]\T$. Furthermore, if $K$ has $d$ non-zero eigenvalues, then the output data points given by $\{y_n^{\rm reduced}\}_{n=1}^N\in\IR^d$ can be found by removing the zero elements in $y_n$.
\end{theorem}

\begin{remark}
   When data is noisy, usually all the eigenvalues of $K$ are non-zero, and therefore, one has to choose the dominant eigenvalues (see \cite{VMS:2008} for some techniques for choosing the dominant eigenvalues). If the eigenvalues of $K$ are sorted in the descending order, the first $d$ elements of $y_n$, $n=1,2,\dots N$, is a $d$-dimensional data set that is \emph{approximately} $k$-locally isometric to $\{x_n\}_{n=1}^{N}\in\IR^D$.
\end{remark}
%%%%%%%%%%%%%%%%%%%%%%%%%%%%%%%%%%%%%%%%%%%%%%%%%%%%%%%%%%%%%%%%%%%%%%%%%%%%%%%
%%%%%%%%%%%%%%%%%%%%%%%%%%%%%%%%%%%%%%%%%%%%%%%%%%%%%%%%%%%%%%%%%%%%%%%%%%%%%%%%

\section{Data Segmentation and Subspace Identification}\label{sec:dataseg}

In this section, we address the problem of data segmentation and subspace identification for a set of given data points. Next, we define the multiple subspace segmentation problem.
\begin{problem}[\hspace*{-.3em}\textbf{(Multiple Subspace Segmentation Problem)}]
  Given the set $Y=\{y_i\}_{i=1}^N$ which are drawn from a set of distinct subspaces of unknown number and dimension, we would like to (\emph{i}) find the number of subspaces, (\emph{ii}) find their dimensions, (\emph{iii}) find the basis for each subspace, and (\emph{iv}) associate each point to the set it belongs to.
\end{problem}

Generalized Principal Component Analysis (GPCA) uses algebraic geometric concepts to address this problem. First, we present the basic GPCA algorithm and later introduce the version of GPCA which is more robust to noise. For a detailed treatment of the subject see \cite{VMS:2008}.

\subsection{Basic GPCA}

In this section we present the Basic GPCA algorithm, where we assume that data points are noise-free. The GPCA algorithm consists of  three main steps: (\emph{i}) polynomial fitting, (\emph{ii}) polynomial differentiation, and (\emph{iii}) polynomial division. Let $\mathcal{A}=\{S_1,S_2,\dots S_n\}$ be a subspace arrangement and $Z_\mathcal{A}=S_1\cup S_2\cup\dots\cup S_n$, where ${\rm dim}(S_j)=d_j$, $j=1,2,\dots n$. Furthermore, let $Y=\{y_i\}_{i=1}^N$ be a set of sufficiently large number of points sampled from $Z_\mathcal{A}$. In this paper, we assume that the number of subspaces $n$ is known. The GPCA algorithm, however, gives the solution for the case where $n$ is unknown (see \cite{VMS:2008}). In order to algebraically represent $\ZA$, we need to find the vanishing ideal of $\ZA$, namely $I(\ZA)$. The vanishing ideal is the set of polynomials which vanish on $\ZA$. It can be shown that the homogenous component of $I(\ZA)$, namely $I_n$, uniquely determines $I(\ZA)$. Therefore, in order to find the vanishing ideal $I(\ZA)$ it suffices to determine the homogenous component $I_n$.

Now note that if $p_n(x)$ is a polynomial in $I_n$ then $p_n(x)=c_n\T \nu_n (x)$, $c_n\in\IR^{M_n(D)}$, where $x=[x_1\, x_2\, \dots \, x_D]\T$, for some $D\in\mathbb{N}$, and $M_n(D)$ is given by (\ref{eq:MnD}). Therefore, each point $y_i$, $i=1,2,\dots N$, should satisfy $p_n(x)$, hence $V_n(D)c_n=0$, where
\begin{eqnarray}
  V_n(D)\defeq\left[
           \begin{array}{c}
             \nu_n\T(y_1) \\
             \nu_n\T(y_2) \\
             \vdots \\
             \nu_n\T(y_N) \\
           \end{array}
         \right]
\end{eqnarray}
is called the embedded data matrix. A one-to-one correspondence between the null space of $V_n(D)$ and the polynomials in $I_n$ exists if the following condition holds
\begin{eqnarray}
  {\rm dim}\left(\mathcal{N}(V_n(D))\right)={\rm dim}(I_n)=h_{\rm I}(n),\label{eq:cond}
\end{eqnarray}
or equivalently,
\begin{eqnarray}
  {\rm rank}\left(V_n(D)\right)=M_n(D)-h_{\rm I}(n),
\end{eqnarray}
where $h_{\rm I}(n)$ is the Hilbert function. The singular vectors of $V_n(D)$ represented by $c_{ni}$, $i=1,2,\dots h_{\rm I}(n)$ corresponding to the zero singular values of $V_n(D)$ can be used to compute a basis for $I_n$, namely
\[I_n={\rm span}\{p_{ni}(x)=c_{ni}\nu_n(x),\, i=1,2,\dots h_{\rm I}(n)\}. \]
In the case where the data $Y$ is corrupted by noise, the singular vectors corresponding to the $h_{\rm I}(n)$ smallest singular values of $V_n(D)$ can be used.

The following theorem shows how polynomial differentiation can be used to find the dimensions and bases of each subspace.

\begin{theorem}[\hspace*{-.3em}\cite{VMS:2008}]
Let $Y=\{y_i\}_{i=1}^N$ be a set of points sampled from $\ZA=S_1\cup S_2\cup\dots\cup S_n$, where $S_i$ is a subspace of unknown dimension $d_i$, $i=1,2,\dots n$. Furthermore, assume that for each subspace $S_j$, $j=1,2,\dots n$, a point $w_j$ is given such that $w_j\in S_j$, $w_j\not\in S_i$, $i\not = j$, $i=1,2,\dots n$, and condition (\ref{eq:cond}) holds. Then
\begin{eqnarray}
  S_j^\perp = {\rm span} \left\{\frac{\partial}{\partial x}c_n\T \nu_n(x)|_{x=w_j}: c_n\in\mathcal{N}\left(V_n(D)\right)\right\},
\end{eqnarray}
where $V_n(D)$ is the embedded data matrix of $Y$. Furthermore, $d_j=D-{\rm rank}\left(\nabla P_n(w_j)\right)$, $j=1,2,\dots n$, where $P_n(x)=[p_{n1}(x)\, p_{n2}(x)\,\dots p_{nh_{\rm I}(n)}(x)]\T \in\IR^{1\times h_{\rm I}(n)}$ is a row vector of independent polynomials in $I_n$ found using the singular vectors corresponding to the zero singular values of $V_n(D)$, and $\nabla P_n=[\nabla\T p_{n1}(x)\, \nabla\T p_{n2}(x)\,\dots \nabla\T p_{nh_{\rm I}(n)}(x)]\in\IR^{D\times h_{\rm I}(n)}$.
\end{theorem}

As a final step, we need a procedure to select a point $w_j$, $j=1,2,\dots n$ for each subspace. Without loss of generality let $j=n$. One can show that the first point $w_n$, where $w_n\in S_n$ and $w_n\not\in S_i$, $i=1,2,\dots n-1$, is given by
\begin{eqnarray}
  w_n={\rm arg min}_{y\in Y:\,\nabla P_n(y)\not = 0} P_n(y)(\nabla\T P_n(y)\nabla P_n(y))^\dag P_n\T (y).
\end{eqnarray}
Furthermore, a basis for $S_n$ can be found by applying PCA to $\nabla P_n(w_n)$. To find the rest of the points $w_i\in S_i$, $i=1,2,\dots n-1$, we can use the polynomial division as proposed by the next theorem.

\begin{theorem}[\hspace*{-.3em}\cite{VMS:2008}]
Let $Y=\{y_i\}_{i=1}^N$ be a set of points sampled from $\ZA=S_1\cup S_2\cup\dots\cup S_n$, where $S_i$ is a subspace of unknown dimension $d_i$, $i=1,2,\dots n$, and suppose (\ref{eq:cond}) holds. Furthermore, let a point $w_n\in S_n$ and $S_n^\perp$ be given. Then the set $\bigcup_{i=1}^{n-1}S_i$ is characterized by the set of homogenous polynomials
\[ \left\{ c_{n-1}\T\nu_{n-1} (x):\, V_n(D)R_n(b_n)c_{n-1}=0,\, \forall b_n\in S_n^\perp,\, c_{n-1}\in\IR^{M_{n-1}(D)}\right\},\]
where $R_n(b_n)\in\IR^{M_n(D)\times M_{n-1}(D)}$ is the matrix of coefficients of $c_{n-1}$ when $(b_n\T x)(c_{n-1}\T\nu_{n-1} (x))\equiv c_n\T \nu_n (x)$ is rearranged to be of the form $R_n(b_n)c_{n-1}=c_n$.
\end{theorem}

Once the homogenous polynomials $\{c_{n-1}\T \nu_{n-1} (x)\}$ given in the previous theorem are obtained, the same procedure can be repeated to find $w_{n-1}$ and the homogenous polynomials characterizing $\bigcup_{i=1}^{n-2} S_i$.

\subsection{Subspace Estimation Using a Voting Scheme}

The Basic GPCA framework works well in the absence of noise. In practice, however, noise is always present and efficient statistical methods need to be used in conjunction with Basic GPCA. In this section, we present one such statistical method where a voting scheme is combined with the Basic GPCA. Here we assume that the number of the subspaces and their dimensions are known. For a more complete treatment of the subject see \cite{VMS:2008}.

Let $Y=\{y_i\}_{i=1}^N\in \IR^D$ be the set of data points sampled from the set $\ZA = S_1\cup S_2\cup\dots\cup S_n$, where $S_j$, $j=1,2,\dots n$, is a subspace of dimension $d_j$ and co-dimension $c_j=D-d_j$. From the discussion in the previous section we know that the homogenous component of degree $n$ of the vanishing ideal $I(\ZA)$ denoted by $I_n$ uniquely defines $I(\ZA)$. Moreover, we mentioned that ${\rm dim} (I_n)=h_{\rm I}(n)$, where $h_{\rm I}(n)$ is the Hilbert function. Let $P=\{p_1(x),\, p_2(x),\,\dots\, p_{h_{\rm I}(n)}(x)\}$ be the set of basis of $I_n$, which can be found by selecting the $h_{\rm I}(n)$ smallest singular values of  $V_n(D)$, where $V_n(D)$ is the embedded data matrix. Suppose we choose a point $y_1\in Y$. Let us define $\nabla P_{\rm B}(y_1)=\left[\nabla\T p_1(y_1)\, \nabla\T p_2(y_1) \, \dots \, \nabla\T p_{h_{\rm I}(n)}(y_1)\right]$.
In the noise-free case ${\rm rank} (\nabla P_{\rm B}(y_1))=c_j$.

However, in the case where the data is corrupted by noise, a more efficient method for computing the bases is desired. Suppose the co-dimension of the subspaces take $m$ distinct values $c_1',c_2',\dots ,c_m'$. In the voting scheme, since we don't know which subspace $y_1$ belongs to and we would like to leave our options open, the base for the orthogonal complement of subspaces of all possible dimensions $c_i'$, $i=1,2,\dots m$, are calculated by choosing the $c_i'$ principal components of $\nabla P_{\rm B}(y_1)$. This results in $m$ matrices $B_i\in\IR^{D\times c_i'}$, $i=1,2,\dots m$ each of which is a candidate base for $S_i^\perp$, $i=1,2,\dots n$.

The idea of the voting scheme is to count the number of repetitions of each candidate base for all points in the data set $y_i$, $i=1,2,\dots N$. At the end, the $n$ bases with the most votes are chosen to be the bases of $S_i^\perp$, $i=1,2,\dots n$, and each point is assigned to the closest subspace. In our criterion for counting the repetition of the bases, two bases are considered to be the same if the angle between the two subspaces spanned by them is less than $\tau$, where $\tau >0$ is a tolerance parameter.

%%%%%%%%%%%%%%%%%%%%%%%%%%%%%%%%%%%%%%%%%%%%%%%%%%%%%%%%%%%%%%%%%%%%%%%%%%%%%%%%%%
\section{Segmentation of Facial Expressions}

In this section, we use the techniques presented in Sections \ref{sec:dimred} and \ref{sec:dataseg} to segment the facial expressions in a given set of images. More specifically, given a set of images of a person with two different facial expressions (e.g. neutral and happy), we try to segment the images based on their facial expression. We should mention that the author in \cite{Turk:1991} uses the idea of point clustering and PCA to segment images with different facial expressions. In this paper, however, we would like to show that if the manifold of faces is ``unfolded'' (e.g. using a Maximum Variance Unfolding technique), different facial expressions reside on different subspaces.

\begin{figure}
\centering
 \psfig{file=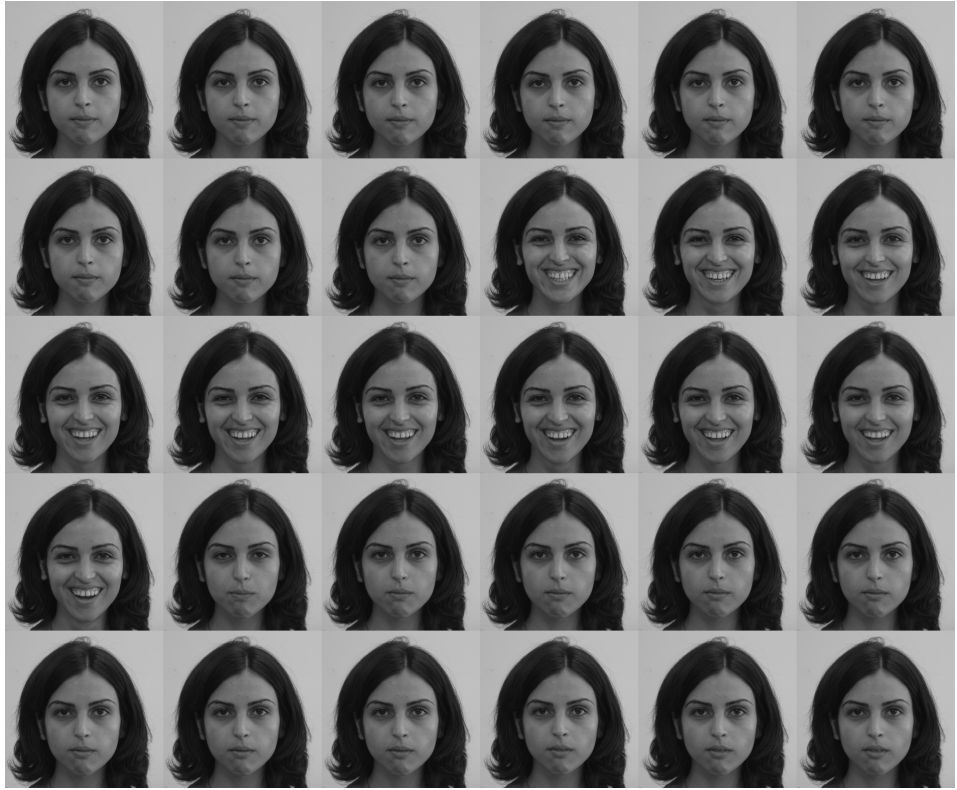,scale = 0.5}
      \caption{A sequence of pictures, where the subject starts with a neutral expression, smiles, and resumes the neutral expression.}
      \label{fig:Negar}
\end{figure}

In our experiment, for each human subject about 30 images of their face were taken, where the subject starts by a neutral expression, transitions to a happy expression, and goes back to the normal expression, where each part contains about 10 images. An example set of images is given in Figure \ref{fig:Negar}. The images were taken in a sequence, each $200\times 240$ pixels, and in total there were 4 subjects.

Each image can be considered as a vector of dimension 48000, by stacking up all the columns in the image matrix. In this way, each image is a point in a 48000 dimension space. In order to segment the images, first, the dimension is reduced to $D=5$ using the MVU procedure presented in Section \ref{sec:dimred}, where $k=4$, i.e. when forming the weighted graph $G$, the 4-nearest neighbors are connected by an edge. Next, the resulting points in the $D=5$ dimensional ambient space are used to identify 2 subspaces of dimension $d=1,2,3,4$, where the in the GPCA voting algorithm two subspace are considered to be the same if the angle between the two is less than $\tau=0.4$ \cite{YRW:ICCV:2005}. The segmentation error for each case is given in Table \ref{tab:tab1}.

\begin{table}
\caption{Segmentation Results for $D=5$} % title name of the table
\centering % centering table
\begin{tabular}{c c cccc} % creating 10 columns
\hline\hline % inserting double-line
Subject & Number of Images &\multicolumn{4}{c}{Segmentation Error}\\ [0.5ex]
 % inserts single-line
 &  &$d=1$ &$d=2$ &$d=3$ &$d=4$\\ \hline
\#1 &29 &3 &2 &2 &3\\
\#2 &31 &13 &13 &3 &7\\
\#3 &31 &6 &15 &2 &4\\
\#4 &32 &13 &15 &1 &1\\
\hline % inserts single-line
\end{tabular}
\label{tab:tab1}
\end{table}

\begin{figure}
\centering
 \psfig{file=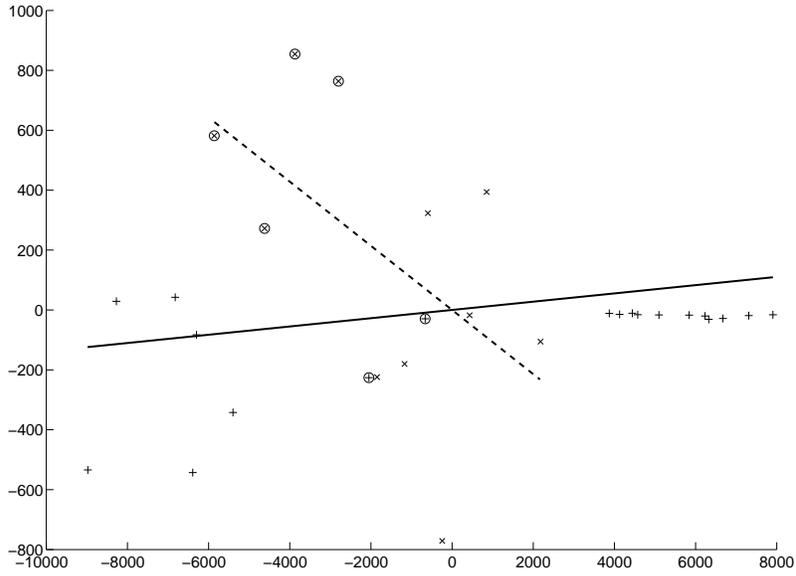,scale = 0.37}
      \caption{Facial expression segmentation with $D=2$ and $d=1$. The categorization error is 6/30. The solid and dashed lines are the subspaces corresponding to the neutral and happy expressions, respectively. The points associated with the solid line and the dashed line are represented by ``$+$'' and ``$\times$'', respectively. The points with ``$\circ$'' are those that are associated with the wrong expression.}
      \label{fig:fig2d}
\end{figure}

 In order to visualize the subspace identification, the segmentation for the case $D=2$, $d=1$ is given in Figure \ref{fig:fig2d}.
\vspace{1em}
%%%%%%%%%%%%%%%%%%%%%%%%%%%%%%%%%%%%%%%%%%%%%%%%%%%%%%%%%%%%%%%%%%%%%%%%%%%%%%%%%
%%%%%%%%%%%%%%%%%%%%%%%%%%%%%%%%%%%%%%%%%%%%%%%%%%%%%%%%%%%%%%%%%%%%%%%%%%%%%%%%%
%%%%%%%%%%%%%%%%%%%%%%%%%%%%%%%%%%%%%%%%%%%%%%%%%%%%%%%%%%%%%%%%%%%%%%%%%%%%%%%%%
\vspace{-1em}
\bibliographystyle{IEEEtran}
\baselineskip=14pt
\bibliography{Reference_new}

\newcommand{\SortNoop}[1]{}
\begin{thebibliography}{1}
\providecommand{\url}[1]{#1}
\csname url@rmstyle\endcsname
\providecommand{\newblock}{\relax}
\providecommand{\bibinfo}[2]{#2}
\providecommand\BIBentrySTDinterwordspacing{\spaceskip=0pt\relax}
\providecommand\BIBentryALTinterwordstretchfactor{4}
\providecommand\BIBentryALTinterwordspacing{\spaceskip=\fontdimen2\font plus
\BIBentryALTinterwordstretchfactor\fontdimen3\font minus
  \fontdimen4\font\relax}
\providecommand\BIBforeignlanguage[2]{{%
\expandafter\ifx\csname l@#1\endcsname\relax
\typeout{** WARNING: IEEEtran.bst: No hyphenation pattern has been}%
\typeout{** loaded for the language `#1'. Using the pattern for}%
\typeout{** the default language instead.}%
\else
\language=\csname l@#1\endcsname
\fi
#2}}

\bibitem{Eisenbud:1995}
D.~Eisenbud, \emph{Commutative Algebra}.\hskip 1em plus 0.5em minus 0.4em\relax
  Springer-Verlag, 1995.

\bibitem{Tannenbaum:1981}
A.~R. Tannenbaum, \emph{Invariance and System Theory: Algebraic and Geometric
  Aspects}.\hskip 1em plus 0.5em minus 0.4em\relax Springer-Verlag, 1981.

\bibitem{Gallian:1990}
J.~A. Gallian, \emph{Contemporary Abstract Algebra}.\hskip 1em plus 0.5em minus
  0.4em\relax D. C. Heath and Company, 1990.

\bibitem{VMS:2008}
R.~Vidal, Y.~Ma, and S.~Sastry, \emph{Generalized Principal Component
  Analysis}.\hskip 1em plus 0.5em minus 0.4em\relax Springer, 2008.

\bibitem{WS:IJCV:2006}
K.~Q. Weinberger and L.~K. Saul, ``Unspurvised learning of image manifolds by
  semi-definite programing,'' \emph{Int. J. Comp. Vis.}, vol.~70, pp. 77--90,
  2006.

\bibitem{Jolliffe:2002}
I.~T. Jolliffe, \emph{Principal Component Analysis}.\hskip 1em plus 0.5em minus
  0.4em\relax Springer, 2002.

\bibitem{TSL:SCI:2000}
J.~B. Tenenbaum, V.~de~Silva, and J.~C. Langford, ``A global geometric
  framework for nonlinear dimensionality reduction,'' \emph{Science}, vol. 290,
  pp. 2319--2323, 2000.

\bibitem{Turk:1991}
M.~A. Turk, \emph{Interactive-Time Vision: Face Recognition as a Visual
  Behavior}.\hskip 1em plus 0.5em minus 0.4em\relax Ph.D Thesis, Massachusetts
  Institute of Technology, 1991.

\bibitem{YRW:ICCV:2005}
A.~Y. Yang, S.~Rao, A.~Wagner, Y.~Ma, and R.~M. Fossum, ``{H}ilbert functions
  and applications to the estimation of subspace arrangements,'' in \emph{Int.
  Conf. Comp. Vis.}, Beijing, China, 2005, pp. 158--165.

\end{thebibliography}
\end{document}